**Tiziana Carpi**
*Department of Studies in Language Mediation and Intercultural Communication*
*University of Milan*

**Stefano Maria Iacus**
*Department of Economics, Management and Quantitative Methods*
*University of Milan*

**Title:** *Is Japanese gendered language used on Twitter ? A large scale study.*

**Abstract**
This study analyzes the usage of Japanese gendered language on Twitter. Starting from a collection of 408 million Japanese tweets from 2015 till 2019 and an additional sample of 2355 manually classified Twitter accounts timelines into gender and categories (politicians, musicians, etc). A large scale textual analysis is performed on this corpus to identify and examine sentence-final particles (SFPs) and first-person pronouns appearing in the texts. It turns out that gendered language is in fact used also on Twitter, in about 6% of the tweets, and that the prescriptive classification into "male" and "female" language does not always meet the expectations, with remarkable exceptions. Further, SFPs and pronouns show increasing or decreasing trends, indicating an evolution of the language used on Twitter.

**Introduction**

Language and gender research studies have described Japanese as a language with particularly marked gender-differentiated forms (Okamoto and Shibamoto, 2016). Despite Japanese speaking practices and Japanese "women's language" (defined as *joseigo*) have been the object of decades of research (Ide 1979, 1997; Shibamoto 1985; Ide, Hori, Kawasaki, Ikuta and Haga 1986; Reynolds and Akiba, 1993), our understanding of Japanese women and men as speaking subjects is still largely influenced by a normative depiction of how they should or are expected to talk. These stereotypical features include different characteristic of speech styles, such as polite, unassertive, empathetic speech, as well as more specific features such as honorifics, self-reference terminologies, sentence-final particles (SFPs), indirect speech acts, exclamatory expressions and voice pitch level (Shibamoto Smith 2003).
Despite these features are assumed to be used only in spoken Japanese, the present study focuses on the analysis of sentence-final particles (SFPs) and personal pronouns as used on Twitter. Contrary to other authors who make use of machine learning to estimate the gender of a Twitter account (see, e.g., Ciot et al., 2013; Sako and Hara, 2013), the aim of this study is about comparing the prescriptive use of SFPs and personal pronouns and their actual use on Twitter. It first focuses on the relative usage of these tokens on a dataset of about 400 millions of tweets collected from late 2015 till early 2019. Regardless of the gender of who actually wrote the tweets, the variation in time is observed and some clear trends emerge. In order to test whether the prescriptive assumption also holds on written Japanese on social media, a subsample of about 2500 accounts has been manually analyzed in order to attribute gender and category (i.e., politician, musician, actor, etc) of the account, by scrutiny of their complete timelines. It emerged that the relative usage of these tokens among males and females subgroups is different and statistically significant confirming the assumption. Further, interesting switching patterns, i.e. "male" markers used by females and viceversa, for some categories are also observed. These analyses have never been conducted so far to the best of authors' knowledge so we think that this paper can pave the way for the construction of new text-based Twitter gender classifiers for Japanese language that, so far, is quite a challenging and unsolved task (see, Ciot *et al.,* 2013). This study also analyzes an unprecedented number of tweets and, by far, uses the most extensive human coded data set of Twitter accounts.

The paper is organized as follows. The next section introduces the reader to the elements of gender markers in Japanese spoken language. Then, the data and the results of the historical analysis are presented. The study of the relative usage of gender markers for testing the prescriptive assumption follows next and finally a discussion closes the paper.

**SENTENCE-FINAL PARTICLES AND FIRST-PERSON PRONOUNS AS GENDER MARKERS IN THE JAPANESE LANGUAGE.**

Bolinger and Sear (1981) define discourse (or pragmatic) markers as *audible gestures* that convey paralinguistic meanings; Japanese sentence-final particles (SFPs) fall into this category. In her introductory textbook on Japanese linguistics, Tsujimura (2002) introduces SFPs as markers that express the speaker's non-propositional modal attitude in casual conversation. In other words, SFPs facilitate the conveyance of the speaker's various emotional states/attitudes in discourse. Yet, Japanese SFPs are also said to be potential - gender - stereotyping sites. (Ide and Yoshida 1999; McGloin 1990; Okamoto 1995; Sturtz and Sreetharan 2004).

According to several studies, in Standard Japanese, the sentence-final particles (SFPs) "わ" *wa* and "かしら" *kashira* have traditionally been associated to as *feminine* speech, "ぜ" *ze* and "ぞ" *zo* as *male* speech and while both are viewed as gender-exclusive, gender-neutral particles like "ね" *ne* or "よ" *yo* have been characterized as mostly *neutral*. Table 1 summarizes SFPs as presented by different authors who have classified them either in three (M, N, F) or five categories: strongly masculine (SM), moderately masculine (MM), neutral (N) strongly feminine (SF), moderately feminine (MF). This classification is a simplified version in which no distinction has been done between strongly and moderately feminine/masculine particles.

| *SFPs* | *Categorization* | SFPs | *Categorization* |
|---|---|---|---|
| ぞ (zo) | **SM** | なの (na no) | [NOM] **SF** |
| ぜ (ze) | **SM** | なのね (na no ne) | [NOM] **SF** |
| さ (sa) | **SM / N** | もん (mon) | **F** |
| な (na) | **N / SM** | もの (mono) | **F** |
| やない (yanai) | **N** | わよ (wa yo) | [VB/ADJ] **SF / SF** |
| やん (yan) | **N** | わね (wa ne) | [VB/ADJ] **SF / SF** |
| やんけ (yanke) | **SM** | わよね (wa yo ne) | [VB/ADJ] **SF / SF** |
| ね (ne) | [VB/ADJ] **N** / [NOM] **SF** | のよ (no yo) | [VB/ADJ] **SF / SF** |
| かい (kai) | **SM** | のね (no ne) | [VB/ADJ] **SF** |
| かな (kana) | **SM** | のよね (no yo ne) | [VB/ADJ] **SF / F** |
| もんな (mon na) | **SM** | だよ (da yo) ∗ | **SM** |
| よな (yo na) | **SM** | だね (da ne) ∗ | **SM** |
| じゃん (jan) | **SM** | だな (da na) ∗ | **SM** |
| け (ke) | **SM** | か (vol. + ka) ∗ | **SM** |
| わ (wa) | **SF ↘ / N ↗** | でしょう (deshou) ∗ | **SF** |
| かしら (kashira) | **SF** | だろう (darou) ∗ | **SF** |
| よ (yo) | **SM / M / N** | でしょう (deshou) ∗ | **SF** |
| よね (yo ne) | [VB/ADJ] **N** | の (no) ∗ | **N ↗** [Q] / [VB/ADJ] **F ↘** |
| なのよ (na no yo) | [NOM] **SF** | | |

**Table 1.** Sentence-final particles. Classification according to different authors. "**SM**" = *strongly masculine*, "**M**" = *masculine*, "**N**" = *neutral*, "**F**" = *feminine*, "**SM**" = *strongly feminine*. '↘' = falling intonation;'↗' = rising intonation; [Q] = at the end of question; [XXX] = preceded by XXX; "∗" correspond to SFPs not used in the analysis.

The particles listed in Table 1 are based on several previous studies on SPFs and on grammar texts (Okamoto, 1992, 1995; Shibamoto Smith, 2003; Okamoto and Shibamoto Smith, 2004; Sturtz Sreetharan 2004; Hiramoto, 2010; Kawasaki and McDougall 2003). Yet the classification is much more complex and diversified than what presented in Table 1 (see, Kawaguchi, 1987; Okamoto and Sato, 1992, for an in depth discussion). It should be noted that Table 1 shows stereotypical gender categorization that is ideological and prescriptive in nature. The SFPs "やん" *yan*, "やない"*yanai*, "やんけ" *yanke* are generally present in the Hanshinskai dialect spoken by people from Kansai area, while all the others are part of the Standard Japanese.

In addition to SFPs, also self-reference features appear to be used in a gender-differentiated way: the repertoire of first-person pronouns of men and women differs as shown in Table 2 and displays 1) a tendency in the use of more formal pronouns by women than by men in referring to themselves, and 2) the near absence of deprecatory/informal expressions for women for the first-person (Ide, 1990).

Examples of gender-exclusive personal pronouns, as prescriptively classified by several authors, are "あたし" *atashi* (feminine), "俺"-"おれ" *ore* and "僕"-"ぼく" *boku* and the second person pronouns "君"-"きみ" *kimi* and "おまえ" *omae* (masculine) (Ide, 1997; Ide and Yoshida, 1999; Shibamoto Smith, 2004).

| *Pronouns* | *Classification* |
|---|---|
| わたくし (watakushi) | **M & F** formal |
| あたくし (atakushi) | **F** formal |
| わたし (watashi) | **M & F** formal / **F** plain |
| わし (washi) | **M** plain |
| ぼく / 僕 (boku) | **M** plain |
| あたし (atashi) | **F** plain/informal |
| おれ / 俺 (ore) | **M** informal |
| あたい (atai) | **F** informal |
| きみ / 君 (kimi) * | **M** informal |
| おまえ (omae) * | **M** informal |

**Table 2.** Classification according to different authors. "*" not used in the analysis. "**M**" = *masculine*, "**F**" = *feminine*, "**M & F**" both genders.

Women's language is a space of discourse that reduces Japanese women to a knowable and unified group, objectifying them through their language use (Inoue, 2006). However, this does not mean that the Japanese express themselves through binary linguistic forms (Nakamura, 2014b). While most studies of Japanese language and gender have tended to focus on these normative usages without examining their implications for the real language practices of real speakers, others have shown that Japanese women and men do not necessarily conform to what they consider to be linguistic gender norms (Okamoto and Shibamoto, 2004; Okamoto and Shibamoto, 2016). It will be interesting to see, as the analysis in the following section will unfold, how these gender markers can be used in practice by both genders.

While in order to provide an exhaustive explanation of SFPs from a semantic and pragmatic point of view would require more space, a few examples (Table 3) will offer an idea of the use and meaning of some of them for the non expert reader.

| a. Iku *zo*. (M)<br>go-Pres | e. Ii desu *ne*. (N)<br>good copula |
|---|---|
| "I am going. I can tell you" | "OK?". |
| b. Sonna koto wa atarimae da *ze*. (M)<br>Such thing obvious copula<br>"That's obvious, I can tell you!" | f. Iku *wa yo ne* (F)<br>go-Pres<br>"I am going, OK?" |
| c. Watashi ga yaru *wa*↑. (F)<br>I do-Pres<br>"I will do it" | g. Iku *no yo ne* (F)<br>go-Pres<br>"I am going, OK?" |
| d. Ii desu *yo*. (N / M)<br>good copula<br>"It's OK". | h. Iku *yo ne* (N)<br>go-Pres<br>"I am going". |

**Table 3.** A few examples of usage of SFPs and first-person pronouns.

According to the above mentioned normative categorization (Table 2), women are supposed to use a higher level of formality, compared to men, when using first-person pronouns in formal contexts (e.g., "わたくし" *watakushi* for women vs "わたし" *watashi* for men), and avoid deprecatory words/expressions to refer to themselves or second person (Ide, 1990). These categorical differences in the repertoire of first-person pronouns, may lead to think that women use automatic expressions of defence and demeanor and are always polite in their speech (Ide, 1990).

Since in Japanese SFPs are elements that mark attitude and/or emotion and therefore occur mostly in speech, with a greater variety found in informal than in formal speech (Narahara, 2005: 151), it may be argued that the current analysis of SFPs in online short written texts (i.e., tweets) is not compatible with their original use in oral communication. However we argue that the distinction between written and conversational discourse, as defined by Reynolds (1985:17), as "highly contrastive in respect to the potential of the addressee's immediate response", has to be reconsidered in the context of Social Network Sites (SNSs). Reynolds explains that in the former, the writer and the reader are spatially separated and coding/decoding is not simultaneous and the writer has no specific reader in mind. At the same time, the reader cannot be expected to participate in the speech act which the writer is performing. In conversation, on the other hand, the speaker is aware of the potential of the listener located within the immediate space of the speaker and the speaker is aware of the potential of the listener's immediate reaction. Most SNSs are now characterized by features that make them look closer to oral communication than written discourse. Nobody can argue today that SNSs do not allow immediate reactions by the potential listeners/readers. Immediacy, on SNSs, is one of the most appreciated features. Twitter, in our case, allows the "speaker" to talk to the universe as well as to an imaginary/ideal "hearer" or, in some cases, even a "real" one.

The aim of the present study is to focus on Japanese as it is used on Twitter, the most popular SNS in Japan with as much as 25.6 million monthly active users in 2019. The main purpose is to investigate how gendered language is used online, and how, considering that Twitter allows us to be anonymous, it may affect the way users (people) express themselves. This will be done by analysing the use of SFPs and first-person pronouns by a very high number of Twitter users as described in the next section. We will use the term *token* to refer to either the *sentence-final particles* (SFPs) and to the *pronouns* listed in Tables 1 and 2 when there is no need to distinguish among the two categories.

**DATA COLLECTION AND RESEARCH METHODOLOGY.**
The data used in this work come from two different repositories. The first larger repository of historical Twitter data used to monitor the usage of each token in time, is part of a collection of Japanese tweets collected through the Twitter search API in the period 24 August 2015 - 22 February 2019 and consisting of 408 Million tweets. This repository was collected with a different scope within the *iGenki* project, a study intended to measure the expressed well-being on social networks, and using only the filter on `language = "Japanese"` and `country = "Japan"`. Twitter search API only provides a 10% sample of all tweets but Twitter does not disclose any information about the representativeness of this sample with respect to the whole universe. Nevertheless, according to our personal experience, also confirmed in Hino and Fahey (2019), the coverage of topics and keywords is quite accurate, so we can consider our repository a sufficiently large and representative sample of the Japanese language "spoken" on Twitter.

From this repository we extracted the tweets containing sentences ending with the 29 SFPs listed in Tables 1 and 8 pronouns as listed in Table 2. We ended up with 26,737,077 tweets (about 6% of the total) written by 1,469,232 unique Twitter accounts.

As we are interested in mapping tokens to gender, and gender information on Twitter is not available most of the times, we extracted a sample of 4,000 accounts and asked human coders to classify them according to the following characteristics: *gender* (male/female/unknown = "cannot guess"/none="no gender, i.e., news outlet, etc") and *type* (individual/group/advertising/brand/media/none of the previous). The human coders have to go through the full Twitter timeline to discern the above two characteristics[1]. In some cases, profiles have been restricted or closed and thus no information could be retrieved.

| male | female | none | unknown | profile closed | Total |
|---|---|---|---|---|---|
| 1,562 | 527 | 85 | 314 | 1,741 | 4,230 |

In addition to the above random selection, a subset of 471 accounts with known gender (254 females, 217 males) and profession (actors, youtuber/blogger, singer, musician, athletes, politicians, cartoonist, tv show participants, etc), was extracted.

In total, we have selected 2,560 accounts, 1,562+217= 1,779 males and 527+254 = 781 females.

For these 2,560 accounts we downloaded the most recent tweets[2] ending up with 2,560,596 different tweets by 2,355 accounts (as for 205 accounts it was not permitted to download the timeline). This set of 2.5 million tweets corresponds to the second set of data that will be analyzed. These tweets span from 2009 to 2020, but we grouped those from 2009 till 2015 together ending up with this distribution of tweets over time:

| <= 2015 | 2016 | 2017 | 2018 | 2019 | 2020 | Total |
|---|---|---|---|---|---|---|
| 364,094 | 333,927 | 327,602 | 350,244 | 548,271 | 636,458 | 2,560,596 |

It should be taken into account that in Japanese the subject is generally omitted and first-person pronouns are used infrequently so, as can be noticed in Figure 2, it should be expected to observe a lower frequency with respect to SFPs.

**RESULTS ON HISTORICAL DATA**

In this section we will focus the analysis on the first set of 26,737,077 historical tweets for which the gender is not known a priori. Figure 1 shows the relative usage over time of each SFPs and pronouns, while Figure 2 shows the absolute frequency. For example, Figure 1 shows that the SFPs "ぞ" zo, "ぜ" ze, "さ" sa, "な" na (row 3, columns 2, 3, 4, 5) show a decreasing trend, as well as "わ" wa (top-left) which was most observed in 2015 (100%) with an evident decrease in use, as low as 40%, in 2020. The first-person pronouns "僕" boku (row 5, column 3) and "わたくし" watakushi instead, display an increasing pattern. Other tokens, e.g., "わし" *washi* (row 2, column 7) have a more stable pattern.

---

[1] This way the Twitter accounts have been classified in this study is human-resources intensive but by far more accurate than simply looking at Twitter accounts name, description and profile, photo as seen in other previous studies.

[2] Up to 3200 tweets are permitted by Twitter API, but in most of the cases we retrieved much less when they correspond to the complete timeline of the accounts.

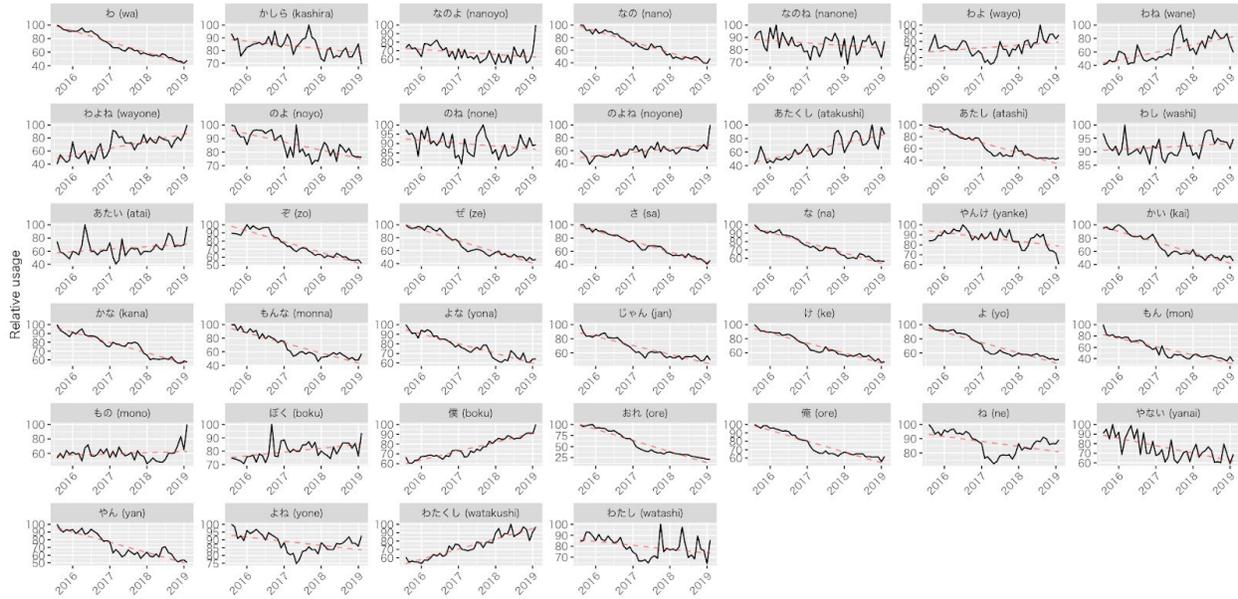

**Figure 1.** The relative usage of each token (SFPs and pronouns) from August 2015 till February 2020. The dashed line represents the trend of the usage with respect to the maximum of the each time series.

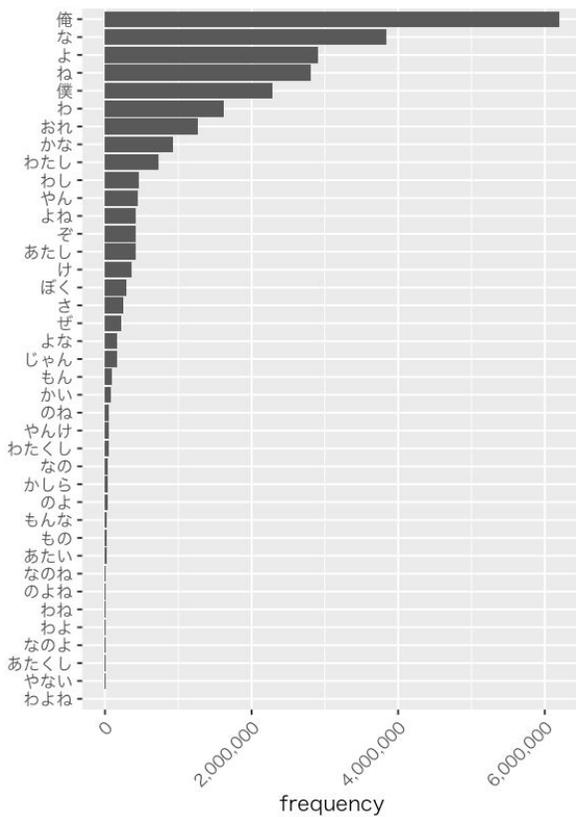
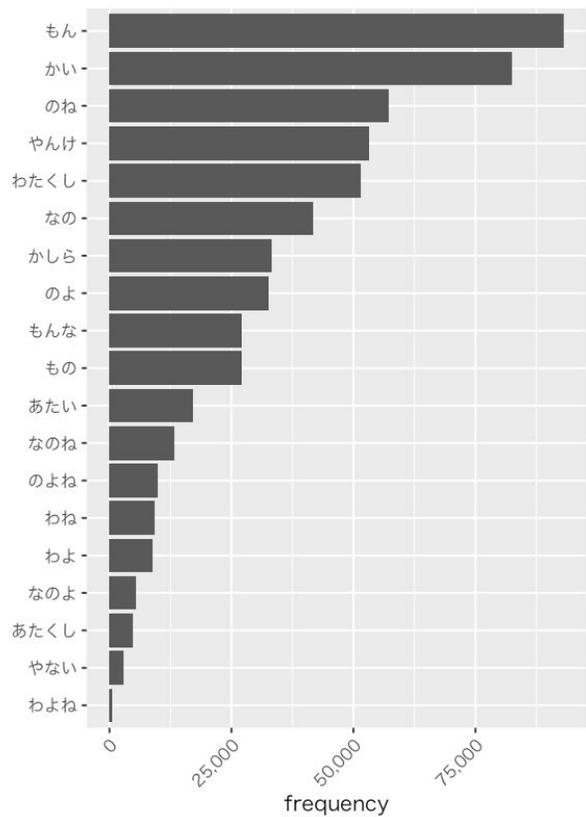

**Figure 2.** The absolute frequencies of counts of each token in our historical dataset. Left panel shows the complete distribution. Right panel is a zoomed plot for tokens that appear less than 100,000 times. The lowest frequency is 540 for "わよね" *wa yo ne*.

**TESTING THE PRESCRIPTIVE ASSUMPTION**

We now focus on the selected sample of 2,560,596 tweets from 2,355 accounts for which human coding identified gender without ambiguity.

To measure the relative importance of the usage of each token in the texts we apply the concept of "text keyness" (see, e.g., Bondi and Scott, 2010) which is essentially a (signed) Pearson's chi-squared test statistics that evaluates a measure of discrepancy between the relative frequencies of each token in the two groups of "males" and "females". Showing the exact value of the test statistics per se is not that important for this analysis, so we use the terms "mostly female" (large positive value of the chi-squared statistics), "almost equal" and "mostly male" (large negative value of the chi-squared statistics) relative usage in the figures.

Figure 3 shows the keyness of each token by gender over time. Lines in the upper part represent tokens mostly used by females, and viceversa. So, for example, the first-person pronouns "わたし" *watashi* and "あたし" *atashi* are more frequent among females, data that confirms what advocated by previous research (respectively as used in formal/plain female speech, and in plain/informal female speech) but the second has a decreasing trend, that is "あたし" *atashi* is used less and less (by female). On the other hand, "俺" *ore* and "僕" *boku* are mostly used by males. The SFP "じゃん" *jan* is equally used by the two groups, i.e., it is not a "key" word. But while, contrary to what has been prescribed by previous studies (see references in the second paragraph of this research) "かしら" *kashira* is, for example, considered a strongly feminine SFP, the results shown in Figure 3 do not display a clear-cut usage: "かしら" *kashira* is surprisingly not employed by women more than it is by men on the analysed tweet dataset. It is interesting to notice that there seems to be some sort of convergence from below and from above towards the centre, that is a tendency for females and male to use mroe neutral forms. Signs of neutralization in the Japanese language had been previously found also by Nakajima (1997) and Takasaki (1997).

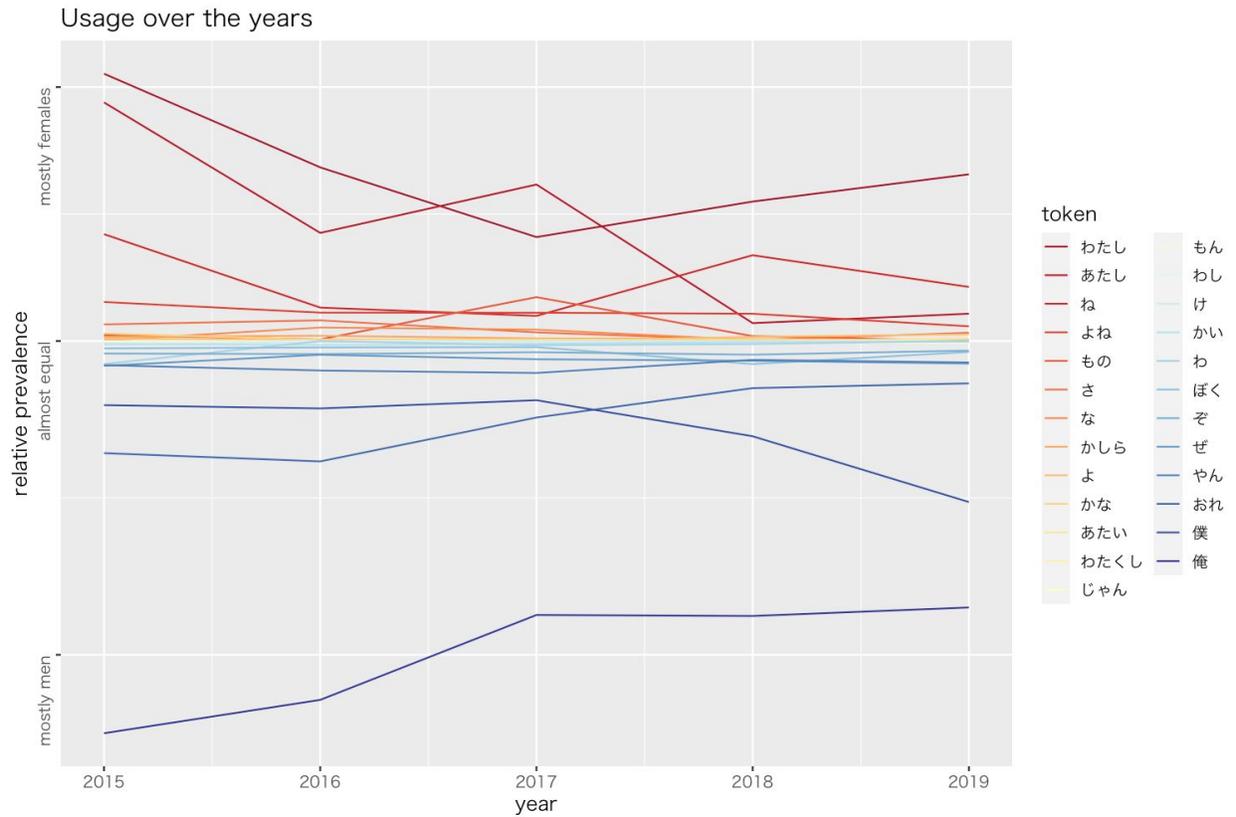

**Figure 3.** Relative usage of token by year. Color represent the prevalent usage by females (reddish) or men (blueish) or almost equal usage (white or faint colors).

Figure 4 shows the overall relative keyness of each token among males and females Twitter accounts. It is clear that "俺" *ore* is almost only used by males and "わたし" *watashi* by females, but not as much. Yet, the length of the bar "わたし *watashi*" is almost half of the one of "俺 *ore*".

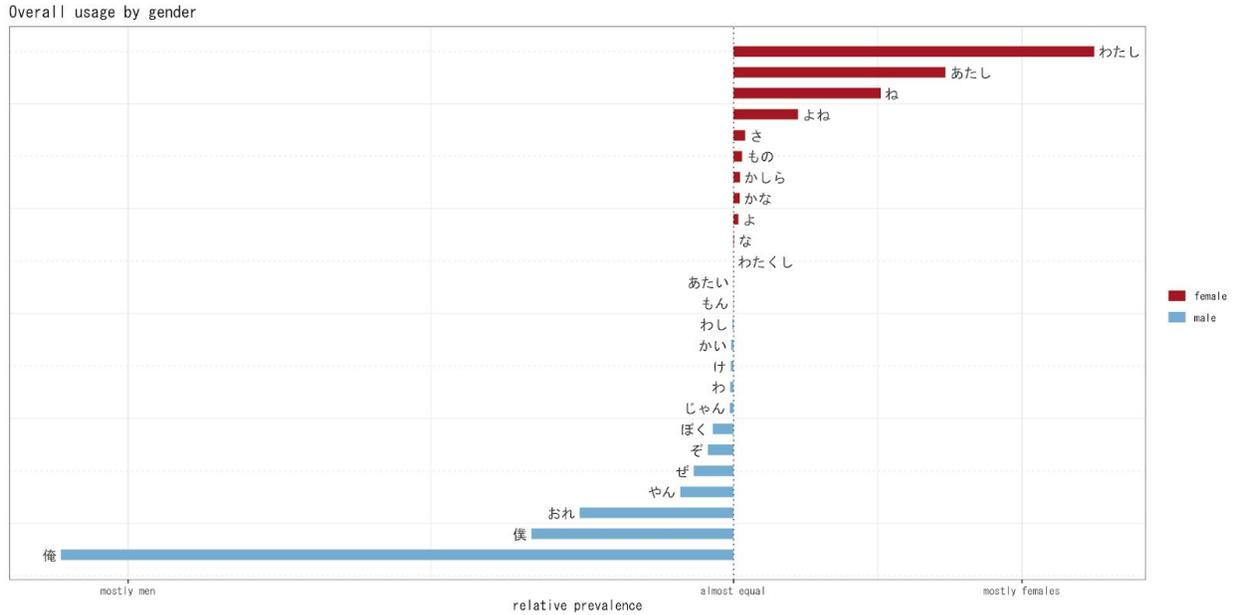

**Figure 4.** Keyness of each token by group irrespective of time.

We then analyzed the same tokens by the groups whose gender has been manually identified. These groups are: *mangaka* (cartoonist, Figure 5), musicians (include singers, Figure 6), actor/actress (and performers in general, Figure 7), politicians (Figure 8), athletes (different types of sport, Figure 9), tv show-related people (anchor persons, talent, tv guest, etc, Figure 10), and YouTuber/blogger (Figure 11). Table 4 summarizes the findings using the following caption: "F➔N" ("M➔N") = slightly more among females (males) but statistically not significant (p-value > 0.05) . "F" / "M" = mostly among females (males), with Chi-squared statistics p-value between 0.01 and 0.05. "FF" / "MM" = almost only among females (males), with Chi-squared statistics p-value less than 0.01 (usually much less). From these results, it is evident that the prescriptive assumption of gender-makers is not always satisfied overall but only for some subgroups (last column of Table 4).

| token | mangaka | actor actress | blogger youtuber | musician | politician | athlete | tv show | overall | Prescriptive assumption | Total tokens |
|---|---|---|---|---|---|---|---|---|---|---|
| わたし | FF | FF | FF | FF | FF | F➔N | FF | FF | M&F formal F plain | 21,406 |
| よ | FF | FF | FF | F➔N | FF | FF | FF | FF | SM/M/N | 516,530 |
| よね | F➔N | FF | FF | FF | FF | F | M➔N | FF | N | 103,853 |
| ね | FF | FF | FF | FF | F➔N | F | FF | FF | N / SF | 484,251 |
| かな | M➔N | FF | FF | FF | M➔N | FF | M➔N | FF | SM | 116,379 |
| かしら | F➔N | FF | F | F | F | FF | F | FF | SF | 8,323 |
| あたい | F➔N | FF | F➔N | F➔N | F➔N | F➔N | F | M➔N | F informal | 620 |
| わ | FF | FF | M➔N | FF | F➔N | FF | F | MM | SF / N | 280,011 |
| あたし | F | F | M➔N | M | -- | FF | FF | FF | F plain F informal | 9,836 |
| わし | M➔N | M➔N | FF | FF | F➔N | F➔N | F➔N | MM | M formal M plain | 6,603 |
| さ | M | MM | F | FF | M➔N | M | F | FF | SM /N | 425,894 |
| な | M | FF | M➔N | FF | MM | M | M➔N | FF | N / SM | 1,202,942 |
| もの | M➔N | F➔N | F➔N | F➔N | F➔N | M➔N | F | FF | F | 71,578 |
| かい | M➔N | M➔N | FF | F➔N | M➔N | F➔N | F➔N | MM | SM | 15,500 |
| もん | M➔N | FF | M➔N | M | M➔N | MM | M➔N | MM | F | 38,507 |
| やん | FF | MM | F➔N | MM | F➔N | M➔N | F➔N | MM | SM | 62,623 |
| わたくし | M➔N | F➔N | MM | MM | M➔N | F➔N | M➔N | F➔N | M & F formal | 1,310 |
| ぜ | M➔N | MM | MM | MM | M➔N | FF | MM | MM | SM | 45,599 |
| け | MM | M➔N | MM | F➔N | FF | MM | M | MM | SM | 181,434 |
| じゃん | MM | F | F➔N | MM | F | M | MM | MM | SM | 25,652 |
| ぞ | MM | M | MM | MM | FF | F | F➔N | MM | SM | 64,622 |
| ぼく | F | F | MM | MM | F➔N | MM | MM | MM | M plain | 9,085 |
| 僕 | MM | MM | MM | MM | MM | MM | MM | MM | M plain | 56,142 |
| 俺 | F➔N | MM | MM | MM | MM | MM | MM | MM | M informal/ deprecatory | 112,680 |
| おれ | MM | MM | MM | MM | F➔N | MM | MM | MM | M informal/ deprecatory | 21,694 |
| # accounts | 49 | 118 | 47 | 94 | 22 | 18 | 46 | 2,355 | | |

**Table 4.** Summary results of relative usage by gender of the different SFPs and personal-pronouns according to different subgroups of accounts and "overall". "F➔N" ("M➔N") = slightly more among females (males) but statistically not significant (p-value > 0.05) . "F" / "M" = mostly among females (males), with Chi-squared statistics p-value between 0.01 and 0.05. "FF" / "MM" = almost only among females (males), with Chi-squared statistics p-value less than 0.01 (usually much less). In the column "Assumption": green color = "assumption verified", red color = "reverse assumption verified" for the overall sample of 2,355 accounts.

Table 4 shows results of the relative usage (by gender) of the different SFPs and first-person pronouns, as they have been used on Twitter by different accounts and in the "overall" column.

The captions, here, show whether the token were mostly used by male (M or MM) or female (F or FF) accounts. The last column displays to which gender the specific token has been associated by previous research on Japanese gendered language, while the color confirms (green) or denies (red) it according to what found in the real conversations/utterances on Twitter. In the case of *"よ" yo, "よね" yone", "かな" kana* and *"わ" wa,* data present a general larger use by females[3] than by males, thus proving the expectations wrong. This means that, for example, *"かな" kana*, usually correlated to men, was found as more used by females on the analyzed tweets. Something similar (but opposite) happens with *"さ" sa* and *"な" na.* However, what is remarkable are those cases in which only one or two subgroups, compared to all the others in the same raw, show a usage by the opposite sex. Clear examples are:

1) *"さ" sa*, usually associated to male usage and confirmed in most columns, was found frequently used also by female musicians, bloggers and tv show-related persons;
2) *"な" na*, usually associated to mostly neutral and male usage, and confirmed in most columns, here results to be employed also by actresses and female musicians;
3) *"け" ke* and *"ぞ" zo*, normatively associated to males, appear to be substantially used also by the opposite sex but only in the case of politicians;
4) "ぜ *ze*, normatively associated to males, appears to be considerably used by female Athletes;
5) "やん" *yan* and "かい" *kai,* two SFPs consistently associated to masculine speech, are found to be strongly used respectively by female *mangaka* (*manga* artists) and by bloggers and Youtubers.

As for first-person pronouns, while the usage of "俺" *ore* (including "おれ, its *hiragana* version)[4] and "僕" *boku* by male users is consistent with what presented by previous research, the following ones present an interesting variation in use in "online speech" on Twitter:

A. *"ぼく" boku* (*hiragana* version of "僕" *boku*)[5] originally associated to male speakers, has been found in tweets by female *mangaka* and actresses.
B. (conversely) *"わたし" watashi* , associated by previous studies to both male and female speech, here is found only among females.

Smith (1992) and Reynolds (1993) write that women in positions of power appear to experience linguistic conflict and that they solve it either by 'defeminizing' their language or creating new strategies to cope with it. These observations are also reported by

---

[3] By "females" and "males" we mean the group of female Twitter accounts and male accounts respectively.
[4] Japanese language is characterized by three different scripts: *kanji*, known as Chinese characters, with semantic values, and the two phonetic syllabaries, *hiragana* (used for suffixes and other grammatical functions) and *katakana* (mainly used for words borrowed from foreign languages). Any Chinese characters can also be written in *hiragana* according to an individual's preference to impart, for example, an informal feel.
[5] See previous note.

Takasaki (1996) on the different speech styles used by females in television interviews. By mixing 'women's language', 'men's language' (*danseigo*) and neutral forms, they enrich their speech and add more expression and colour to their account (Takasaki, 1996).

Our analysis suggest that women politician may want to use "け" *ke* and "ぞ" *zo* in order to reject the dominant gender and sexual norms, empowering their speech while negotiating their identity and power. Something similar has been found in the use of first-person pronouns by junior high-school female students who used to negotiate their power through the use of the normative masculine "僕" *boku* (Miyazaki, 2004). As for the unexpected use of "やん" *yan* and "かい" *kai*, and of "ぼく" *boku* by *mangaka*, the reasons may be more complex and be related also to the use of "role language" (*yakuwarigo*)[6] that the current research does not cover.

Such complex, contradictory, and unexpected processes can be understood only by following the naturally occurring linguistic practices of a specific community as has been done in this study on Twitter. However, only a close examination of language use in real contexts would enable us to fully understand the complicated and dynamic relationship between specific linguistics choices and their social meaning.

The present study is mostly descriptive as it takes into account only the presence of SFPs and first-person pronouns. A more extensive semantic analysis will be the object of further investigation.

**CONCLUDING REMARKS**

This work presented the first large scale study of Japanese gendered language on Twitter on both historical data (408 million tweets from 2015 till 2019), and a large sample of 2,355 manually scrutinized Twitter accounts.

This study has shown that the real use of gendered language on a SNS such as Twitter does not always meet the expectations according to ideological linguistic norms thus reinforcing what stated by Okamoto and Shibamoto Smith (2016: 22): "while Japanese linguistic gender norms may appear to be firmly established in society, closer examination reveal highly diverse views about and uses of gendered speech, suggesting that normative meanings are contested and that the indexical fields of gendered speech forms are potentially broader and variable".

---

[6] Sets of spoken language features (vocabulary and grammar) and phonetic characteristics (intonation and accent patterns) psychologically associated with particular character types are termed "role language" (*yakuwarigo*) (Kinsui, 2003).

The analysis also revealed that the use of gendered language changes through time and also across different subgroups of Twitter accounts, indicating either a strategic (politicians) or a further stereotyped (*mangaka*) usage.

The use of gendered language on SNS is a relatively unexplored area in the field of sociolinguistics and there are few studies on the subject. The authors hope that this work will contribute to raise interest and inspire further studies on the subject.

This study also paves the way for new machine learning Twitter accounts classifiers.

**ACKNOWLEDGMENTS**

The data collection has been performed within the Japan Science and Technology Agency CREST (Core Research for Evolutional Science and Technology) project, grant n. JPMJCR14D7. We also acknowledge Marina Gaeta, Lucia Mancini, Chiara Ricupero and Chen Xing Yi for their accurate coding activity and participation to the project.

Nakajima, E. (1997). *Gimon hyōgen no yōsō* [An aspect of interrogative expressions]. In *Gendai Nihongo Kenkyūkai*. [Modern Japanese Research Group] Josei no Kotoba. Shokubahen [Women and Language. At the workplace], 59–82. Tokyo: Hitsuji Shobo

Nakamura, M.. (2014a). "Historical Discourse Approach to Japanese Women's Language: Ideology, Indexicality, and Metalanguage." In The *Handbook of Language, Gender, and Sex- uality*, 2nd ed, ed. by Susan Ehrlich, Miriam Meyerhoff and Janet Holmes, Oxford: Wiley-Blackwell, pp. 378–395.

Nakamura, M. (2014b). *Gender, Language and Ideology: A genealogy of Japanese women's language.* John Benjamin's Publishing Company.

Okamoto, S. (1995). "Tasteless" Japanese: Less "Feminine" Speech among Young Japanese Women. In Hall, K. and Bucholtz, M. (1995) *Gender Articulated: Language and the Socially Constructed Self* (pp. 297-325), New York: Routledge.

Okamoto, S., Shibamoto Smith, J.S. (2016) *The Social Life of the Japanese Language: Cultural Discourse and Situated Practice*, Cambridge University Press.

Okamoto, S., Shibamoto Smith, J.S. (2004) *Japanese Language, Gender, and Ideology: Cultural Models and Real People*. Oxford, UK: Oxford University Press.

Okamoto, S., Sato, S. (1992). Less feminine speech among young Japanese females. In K. Hall, M. Bucholtz, and B. Moonwomon (eds.), *Locating power: Proceedings of the Second Berkeley Women and Language Conference*, vol. 2, 478–488. Berke- ley: Berkeley Women and Language Group.

Reinoruzu [Reynolds] A., K. (ed.) (1993). *Onna to nihongo* (*Women and the Japanese language*). Tokyo: Yushindo.

Reynolds, K. A. (1990) Female speakers of Japanese in transition. In Sachiko Ide and Naomi Hanaoka McGloin (ed.), *Aspects of Japanese women's language*, pp. 129-146. Tokyo: Kuroshio Shuppan.

Reynolds, K. A. (1985). Female speakers of Japanese. *Feminist Issues* 5: 13–46.

Shibamoto, J. S. (1985). *Japanese women's language*. New York FL: Academic Press.